\documentclass[11pt]{article}

\ifx\review\undefined
\usepackage[final]{acl}
\else
\usepackage[review]{acl}
\fi

\usepackage{times}
\usepackage{hyperref}
\usepackage{amsmath}
\usepackage{amssymb}
\usepackage[T1]{fontenc}
\usepackage[utf8]{inputenc}
\usepackage{microtype}
\usepackage{inconsolata}
\usepackage{graphicx}
\usepackage{float}
\usepackage[belowskip=-0.8em,labelfont=bf,labelsep=endash]{caption}
\usepackage{xcolor}
\usepackage{sourcecodepro} %
\usepackage{listings}
\usepackage{booktabs}
\usepackage{longtable}
\usepackage{fontawesome}
\usepackage{scalerel}
\usepackage{makecell}
\usepackage{enumitem}
\usepackage{tabularx}
\usepackage{subcaption}
\usepackage{stfloats}
\usepackage{tikz}
\usetikzlibrary{arrows.meta, positioning, matrix, fit, backgrounds, calc}

\usepackage{mathtools}
\usepackage{amsthm}

\usepackage[capitalize,noabbrev]{cleveref}

\renewcommand{\sectionautorefname}{\S\kern-3pt}
\renewcommand{\appendixautorefname}{\S\kern-3pt}

\definecolor{panelBg}{HTML}{F8F6F0}
\definecolor{panelBorder}{HTML}{DDDDDD}
\definecolor{textColor}{HTML}{111111}

\definecolor{qqBlue}{HTML}{1F4B99}
\definecolor{qqGreen}{HTML}{009B55}
\definecolor{qqGreenText}{HTML}{2F6B3B}
\definecolor{qqOrange}{HTML}{C04F17}
\definecolor{qqOrangeText}{HTML}{8B5A2B}
\definecolor{qqMuted}{HTML}{666666}
\definecolor{qqMutedSoft}{HTML}{A7A29B}
\definecolor{qqBackground}{HTML}{FBFBF8}
\definecolor{qqBackgroundTop}{HTML}{F6F4EF}
\definecolor{qqPanel}{HTML}{FFFFFF}

\definecolor{charcoal}{HTML}{2D2D2D}
\definecolor{cleanbackground}{HTML}{F6F6F6}
\definecolor{mutedgreen}{HTML}{6A9955}
\definecolor{mutedred}{HTML}{A4544A}
\definecolor{mutedorange}{HTML}{C56F15}
\definecolor{softgold}{HTML}{C1A070}
\definecolor{functionblue}{HTML}{4A78A0}
\definecolor{commentgray}{HTML}{75715E}
\lstdefinestyle{cleanpython}{
    language=Python,
    backgroundcolor=\color{cleanbackground},
    basicstyle=\ttfamily\small\color{charcoal}, 
    frame=single,
    framesep=6pt, %
    framerule=0pt, %
    stringstyle=\color{mutedgreen}, %
    keywordstyle=\color{charcoal}\bfseries, %
    commentstyle=\color{commentgray}\itshape, %
    identifierstyle=\color{charcoal},
    showtabs=false,
    showspaces=false,
    showstringspaces=false,
    breaklines=true,
    breakatwhitespace=true,
    morekeywords={self, class, def, return, True, False, None, lambda, yield, from, import, assert},
    inputencoding=utf8,
    extendedchars=true,
    literate={á}{{\'a}}1 {ã}{{\~a}}1 {é}{{\'e}}1, %
}

\definecolor{SubtleBlue}{HTML}{2C5F80}

\newcommand{\qwanqwa}{%
    \raisebox{-0.3\baselineskip}{%
        \includegraphics[height=1.2em]{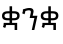}%
    }%
}

\newcommand{\yes}{\textcolor{qqGreen}{\faCheck}} 
\newcommand{\notsure}{\textcolor{qqMuted}{\faQuestion}}
\newcommand{\partially}{\textcolor{qqBlue}{\faCircleO}}
\newcommand{\no}{\textcolor{qqOrange}{\faClose}}

\newcommand{\faGitLogo}{\scalebox{1.2}{\scalerel*{\includegraphics{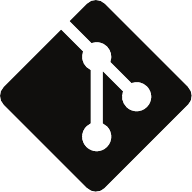}}{X}}}

\newcommand{\enumsepsize}{0pt}

\newcommand{\lagom}{L\textsuperscript{\hspace{-3pt}A}G\textsubscript{O}M$\boldsymbol{\cdot}$NLP}

 \author{Wessel Poelman${}^1$ \and Yiyi Chen${}^2$ \and Miryam de Lhoneux${}^1$\\
  ${}^1$\lagom, Department of Computer Science, KU Leuven\\
  ${}^2$AAU-NLP, Department of Computer Science, Aalborg University\\
  \texttt{wessel.poelman@kuleuven.be}
}

\title{QQ: A Language Metadata Toolkit for Multilingual NLP}

\begin{document}
\maketitle

\begin{abstract}
    Multilingual NLP research increasingly involves hundreds or thousands of languages across different datasets.
    Managing, discovering, and reporting language metadata becomes a common hurdle at these scales.
    We present QQ, a metadata toolkit and browser explorer. 
    QQ compiles language metadata sources into a graph of language varieties, scripts, regions, identifiers, names, and relations, and exposes it through a Python API, a command-line interface, and a browser-based explorer.
    Users can normalize identifiers, retrieve metadata, traverse relations, and discover which external resources contain which languages.
    We demonstrate QQ on three workflows: an audit of the HuggingFace Hub, linking resources that use different identifier systems, and generating language-reporting tables.
    QQ supports FAIR-oriented metadata practices through versioning, open formats, and reusable interfaces.
\end{abstract}

\begin{center}\large
    \begin{tabular}{ll}
    \faGlobe\ \href{https://wesselpoelman.nl/qq/}{Explorer} &
    \faGitLogo\ \href{https://github.com/WPoelman/qwanqwa}{Source}\\
    \faTerminal\ \href{https://pypi.org/project/qwanqwa/}{PyPI} & \faYoutube\ \href{https://youtu.be/RMTqWq4YMZw}{Demo}
    \end{tabular}
\end{center}

\section{Introduction}
The number of languages considered in multilingual NLP research has increased drastically in recent years \cite[\emph{e.g.,}][]{adebara2023serengeti,imanigooghari2023glot500}.
Datasets, benchmarks, and models now regularly cover hundreds or thousands of languages.
While good news for the field, this also creates a practical problem: language metadata becomes harder to manage consistently.

Additionally, multilingual models are increasingly evaluated on more tasks and in more languages.
As more datasets are combined across training and evaluation, a common problem becomes more apparent: datasets do not stick to one convention when using language identifiers.
One dataset might list German as \texttt{de}, another as \texttt{deu}, and a third as \texttt{stan1295}.
This is easy to handle when dealing with ten or twenty languages, but with hundreds or thousands of languages, with several identifier types in active use, it becomes cumbersome at best and unmanageable at worst.
This issue is amplified since languages need to be tracked across exploration, data cleanup, analysis, and reporting.

Identifiers are only part of the problem.
Language metadata matters \emph{a posteriori} when reporting which languages were used in a study \citep{bender2019benderrule}, and it matters \emph{a priori} when choosing languages based on scripts, geography, families, or other properties.
Currently, managing this often means using local mapping files, scattered scripts, and manual browsing across several metadata sources, such as Wikipedia or Glottolog \cite{hammarstrom2024glottolog}.
As the number of languages grows, one-off scripts become increasingly brittle: the same identifiers must be resolved repeatedly across different resources and experimental steps.

Coming back to our German example:
OPUS \citep{tiedemann2012parallel} and Universal Dependencies \citep{nivre2020universal} use the ISO 639-1 code \texttt{de}; WMT uses the ISO 639-3 code \texttt{deu}; older datasets might use ISO 639-2 \texttt{ger}; some datasets, such as NLLB \citep{costa-jussa2024scaling} and FineWeb-2 \citep{penedo2025fineweb2}, use tags such as \texttt{deu\_Latn}; linguistic datasets often use the Glottocode \texttt{stan1295} \citep{forkel2022glottocodes}; and Wikidata uses \texttt{Q188}.
These different identifiers all refer to the same language.

The same problem appears when trying to find datasets for a particular language.
On the HuggingFace Hub \cite{lhoest2021datasets}, the language filter shows separate entries for what is effectively the same language, such as \texttt{el} and \texttt{ell} for Modern Greek.
Some commonly-used linguistic resources, like WALS \cite{wals}, invent their own identifiers that do not belong to any other standard (\texttt{grk} for Modern Greek). 
This makes the discoverability of which datasets contain which languages more difficult than it should be.
This issue is only getting worse given the increasing number of languages and datasets in multilingual NLP research.
Routine tasks like finding datasets for a language, counting language coverage, or combining datasets should be quick and convenient.

To address these issues, we introduce QQ.
\textbf{Q}wan\textbf{Q}wa is the phonetic spelling of \qwanqwa, meaning \emph{language} in Amharic.
QQ compiles language metadata sources into a versioned, unified database. 
This database serves as a shared backend for three complementary interfaces: a Python library for simple, programmatic \emph{management}, a command-line interface for simple, \emph{reproducible} workflows, and a browser explorer for interactive \emph{discovery}.
QQ resolves identifiers, merges sources, and allows the same data to be reused in scripts, reports, and discovery.
Our contributions are:
\begin{itemize}[itemsep=\enumsepsize,topsep=2pt]
    \item We introduce QQ, a language metadata toolkit and browser explorer for multilingual NLP.
    \item QQ integrates open language metadata resources into one graph over language varieties, scripts, regions, families, identifiers, names, speaker estimates, endangerment statuses, dataset availability, and more.
    \item We validate QQ on practical workflows: large-scale language-tag auditing, resource linking, and language reporting.
    \item QQ supports FAIR-oriented metadata practices \cite[\emph{Findable, Accessible, Interoperable, Reusable;}][]{wilkinson2016fair} through versioning, open formats, and reusable interfaces.
\end{itemize}

\section{QQ}
The core of QQ is a compiled database with several interfaces on top.
During database construction, QQ imports open language metadata sources, normalizes and resolves their identifiers, and merges partially overlapping records into \textit{a unified metadata graph}.
The compiled database ships with the Python package, so common workflows work immediately after installation without downloading or querying each source separately.
This also ensures proper versioning and reproducibility.
The same snapshot is then used by the Python API, command-line interface, and browser explorer.

QQ represents metadata as a graph consisting of \texttt{Languoid}, \texttt{Script}, and \texttt{Region} nodes.
We use ``languoid'' as a neutral term for language-like entities covering languages, dialects, macrolanguages, and families \citep{good2006modeling}.
Identifiers do not always refer only to individual languages: ISO 639-3 codes and Glottocodes can refer to all four, whereas ISO 639-5 codes strictly refer to families, for example.
Edges in the graph connect languoids to scripts, regions, and other languoids.
Other metadata such as level (language, dialect, family), identifiers, names, speaker counts, endangerment, external links, and more are stored as node or edge properties.

\begin{figure*}[t]
    \centering
    \begin{tikzpicture}[
    >=Stealth,
    font=\rmfamily,
    panel/.style={
        draw=panelBorder,
        fill=qqBackgroundTop,
        rounded corners=6pt,
        minimum width=4.05cm,
        minimum height=5.9cm,
        line width=0.8pt
    },
    header/.style={
        font=\small\bfseries,
        text=textColor,
        text width=3.55cm,
        align=center,
        anchor=north
    },
    neutralbox/.style={
        draw=panelBorder,
        fill=qqPanel,
        rounded corners=4pt,
        inner xsep=6pt,
        inner ysep=5pt,
        text width=3.12cm,
        align=center,
        font=\scriptsize,
        line width=0.7pt,
        text=textColor
    },
    languoid/.style={
        draw=qqBlue,
        fill=qqBlue!10,
        rounded corners=4pt,
        inner sep=7pt,
        text width=3.05cm,
        align=center,
        line width=0.8pt,
        text=textColor
    },
    entity/.style={
        draw=#1,
        fill=#1!10,
        rounded corners=4pt,
        inner xsep=7pt,
        inner ysep=5pt,
        minimum width=1.43cm,
        align=center,
        font=\scriptsize,
        line width=0.8pt,
        text=textColor
    },
    route/.style={draw=qqMuted, line width=0.6pt, line cap=round, line join=round},
    arrow/.style={->, draw=qqMuted, line width=0.65pt, line cap=round, line join=round, shorten >=2.5pt},
    graphEdge/.style={<->, draw=qqMuted, line width=0.6pt, line cap=round, shorten >=2pt, shorten <=2pt},
    continuation/.style={
        -{Stealth[length=3pt,width=3pt]},
        draw=qqMutedSoft,
        densely dashed,
        opacity=0.55,
        line width=0.5pt,
        line cap=round,
        shorten >=1pt
    }
]

    \node[panel] (panel1) at (0,0) {};
    \node[panel, right=0.85cm of panel1] (panel2) {};
    \node[panel, right=0.85cm of panel2] (panel3) {};

    \node[header, yshift=-5pt] (head1) at (panel1.north) {1.\enspace Metadata sources};
    \node[header, yshift=-5pt] (head2) at (panel2.north) {2.\enspace Resolve into QQ};
    \node[header, yshift=-5pt] (head3) at (panel3.north) {3.\enspace Research uses};
    \draw[panelBorder, line width=0.6pt]
        ([xshift=9pt,yshift=-24pt]panel1.north west) --
        ([xshift=-9pt,yshift=-24pt]panel1.north east);
    \draw[panelBorder, line width=0.6pt]
        ([xshift=9pt,yshift=-24pt]panel2.north west) --
        ([xshift=-9pt,yshift=-24pt]panel2.north east);
    \draw[panelBorder, line width=0.6pt]
        ([xshift=9pt,yshift=-24pt]panel3.north west) --
        ([xshift=-9pt,yshift=-24pt]panel3.north east);

    \node[neutralbox, anchor=north, yshift=-34pt] (res1) at (panel1.north) {
        \textbf{Resource 1}\\[2pt] \texttt{de}\quad \texttt{deu}
    };
    \node[neutralbox, below=6pt of res1] (res2) {
        \textbf{Resource 2}\\[2pt] \texttt{deu}\quad \texttt{stan1295}
    };
    \node[neutralbox, below=6pt of res2] (res3) {
        \textbf{Resource 3}\\[2pt] \texttt{Q188}\quad \texttt{de}\quad \texttt{ger}
    };
    \node[neutralbox, below=6pt of res3] (resN) {
        \textbf{Resource $N$}\\[2pt] \ldots
    };

    \node[languoid, anchor=north, yshift=-34pt] (merged) at (panel2.north) {
        {\small\bfseries\color{qqBlue}Languoid:}\enspace{\small\bfseries German}\\[5pt]
        \begin{tabular}{@{}l@{\hspace{8pt}}l@{}}
            {\scriptsize\color{qqMuted}ISO 639} & {\scriptsize\ttfamily de\enspace ger\enspace deu}\\
            {\scriptsize\color{qqMuted}Glottocode} & {\scriptsize\ttfamily stan1295}\\
            {\scriptsize\color{qqMuted}Wikidata} & {\scriptsize\ttfamily Q188}
        \end{tabular}
    };
    \node[entity=qqOrange, below=25pt of merged.south, xshift=-0.83cm] (script) {
        {\bfseries\color{qqOrangeText}Script}\\[1pt]Latin
    };
    \node[entity=qqGreen, below=25pt of merged.south, xshift=0.83cm] (region) {
        {\bfseries\color{qqGreenText}Region}\\[1pt]Germany
    };
    \draw[graphEdge]
        ([xshift=-0.48cm]merged.south) -- (script.north);
    \draw[graphEdge]
        ([xshift=0.48cm]merged.south) -- (region.north);

    \draw[continuation]
        ([xshift=-0.16cm]script.north) -- ++(-0.19cm,0.42cm);
    \draw[continuation]
        ([xshift=0.16cm]region.north) -- ++(0.19cm,0.42cm);
    \draw[continuation]
        (script.west) -- ([xshift=3pt]panel2.west |- script.west);
    \draw[continuation]
        ([xshift=-0.16cm]script.south) --
        ([xshift=-1.18cm,yshift=3pt]panel2.south);
    \draw[continuation]
        ([xshift=0.16cm]script.south) --
        ([xshift=-0.48cm,yshift=3pt]panel2.south);
    \draw[continuation]
        (region.east) -- ([xshift=-3pt]panel2.east |- region.east);
    \draw[continuation]
        ([xshift=-0.16cm]region.south) --
        ([xshift=0.48cm,yshift=3pt]panel2.south);
    \draw[continuation]
        ([xshift=0.16cm]region.south) --
        ([xshift=1.18cm,yshift=3pt]panel2.south);

    \node[neutralbox, anchor=north, yshift=-34pt] (use1) at (panel3.north) {
        \textbf{Linking}\\[2pt] merge datasets through shared identifiers
    };
    \node[neutralbox, below=12pt of use1] (use2) {
        \textbf{Management}\\[2pt] normalize and enrich metadata
    };
    \node[neutralbox, below=12pt of use2] (use3) {
        \textbf{Discoverability}\\[2pt] find languoids, scripts, regions, and linked resources
    };

    \coordinate (mergeBus) at ($(panel1.east)!0.5!(panel2.west)$);
    \draw[route] (mergeBus |- res1.east) -- (mergeBus |- resN.east);
    \draw[route] (res1.east) -- (mergeBus |- res1.east);
    \draw[route] (res2.east) -- (mergeBus |- res2.east);
    \draw[route] (res3.east) -- (mergeBus |- res3.east);
    \draw[route] (resN.east) -- (mergeBus |- resN.east);
    \draw[arrow] (mergeBus |- merged.west) -- (merged.west);

    \coordinate (branchBus) at ($(panel2.east)!0.5!(panel3.west)$);
    \draw[route] (merged.east) -- (branchBus |- merged.east);
    \draw[route] (branchBus |- use1.west) -- (branchBus |- use3.west);
    \draw[arrow] (branchBus |- use1.west) -- (use1.west);
    \draw[arrow] (branchBus |- use2.west) -- (use2.west);
    \draw[arrow] (branchBus |- use3.west) -- (use3.west);

\end{tikzpicture}
    \caption{QQ links metadata sources through shared identifiers. Once records are resolved to the same languoid, the merged representation supports normalization, linking, traversal, reporting, and resource discovery.}
    \label{fig:identifier-linking}
\end{figure*}

\begin{figure*}[t]
    \centering
    \includegraphics[width=0.94\linewidth,trim=0 170 0 0,clip]{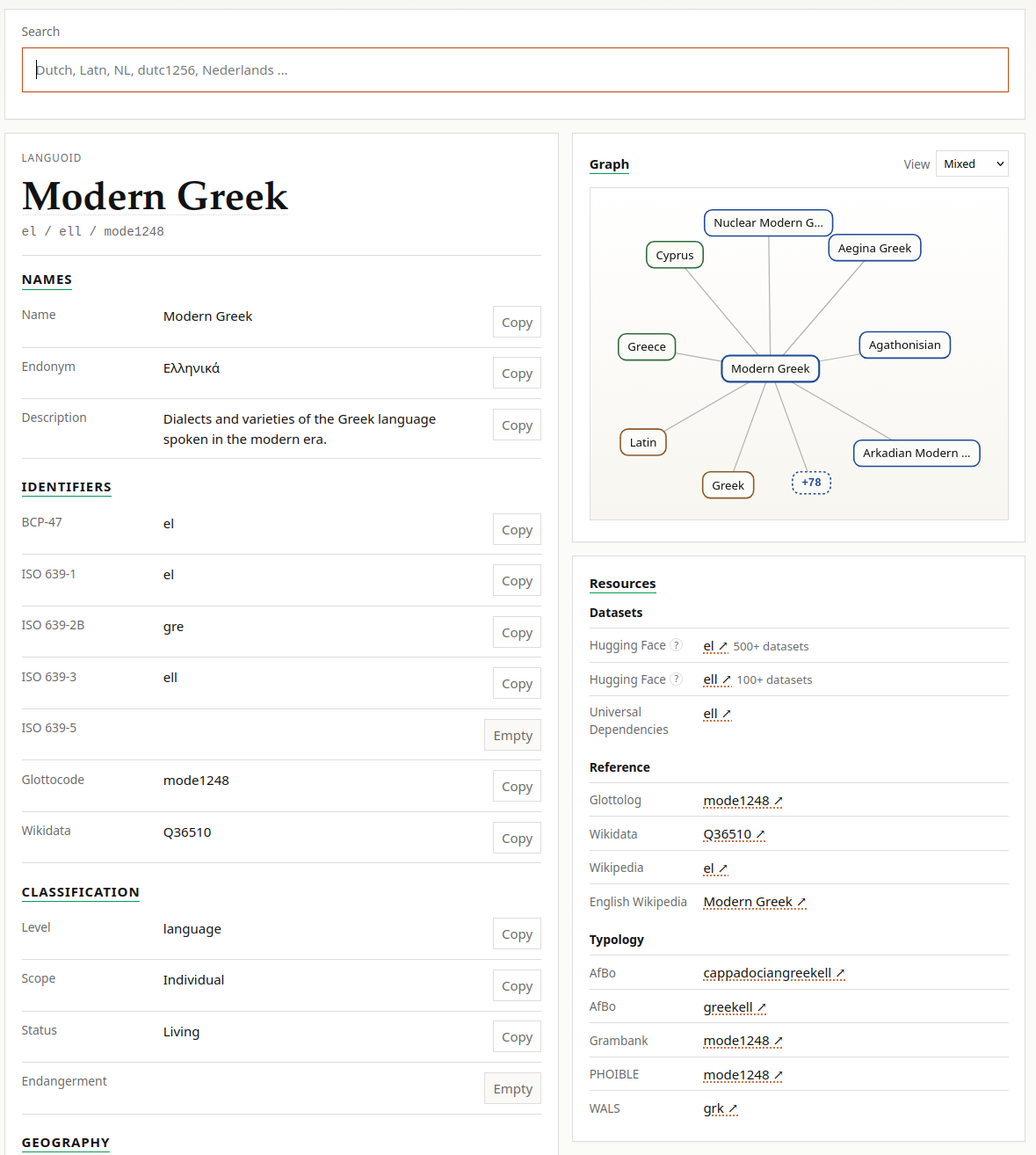}
    \caption{QQ browser explorer showing identifiers, metadata, graph navigation, and resources for Modern~Greek.}
    \label{fig:browser}
\end{figure*}

QQ focuses on sources useful for multilingual NLP workflows.
It currently integrates LinguaMeta \citep{ritchie2024linguameta}, Glottolog \citep{hammarstrom2024glottolog}, Glotscript \citep{kargaran2024glotscript}, Wikipedia and Wikidata metadata, ISO data from the Library of Congress,\footnote{\href{https://www.loc.gov/librarians/standards}{loc.gov/librarians/standards}} script information from Unicode,\footnote{\href{https://www.unicode.org/ucd/}{unicode.org/ucd}} and deprecated-code information from SIL\footnote{\href{https://iso639-3.sil.org/}{iso639-3.sil.org}} and IANA.\footnote{\href{https://www.iana.org/protocols}{iana.org/protocols}}
These sources provide overlapping but different information: identifiers, language names, speaker counts, family relations, scripts, regions, Wikipedia editions, and deprecated or replacement identifier codes.
The merge step uses shared identifiers to connect partially overlapping source records, as illustrated in \autoref{fig:identifier-linking}.
After merging, QQ can resolve all identifiers in the connected record to the same languoid.
These resources are the \emph{primary} sources for the graph; QQ also supports linking to \emph{external} sources (as seen in \autoref{fig:browser}).
This allows for easy discoverability of resources linked to a particular language, regardless of the identifier convention it uses.
Full source details are available in \autoref{app:sources}.

In Python, QQ supports many common operations through a compact interface:\footnote{See the \href{https://github.com/WPoelman/qwanqwa}{README} for much more functionality.}

\begin{lstlisting}[style=cleanpython]
from qq import Database, IdType
db = Database.load()

d1 = db.get("nl")
d1.name 
> "Dutch"
d1.iso_639_3 
> "nld"
d1.speaker_count 
> 24085200
d1.parent.parent 
> Languoid("Modern Dutch")

d3 = db.guess("dutc1256")
d2 = db.guess("nld_Latn")
d4 = db.guess("nl-Latn-NL")
d1 == d2 == d3 == d4 # True
\end{lstlisting}

Three design goals were especially important.
First, QQ ships a packaged snapshot,\footnote{The database is $\sim$3 MiB; lazy-loaded name data $\sim$8 MiB.} so common workflows work immediately after installation.
Second, identifier resolving is the central operation, which means the Python API, CLI, browser explorer all use the same database.
Third, the graph structure: multilingual NLP workflows often involves multi-hop traversal: from a language to a region, from that region to other languages, and from languages to scripts or linked resources. 
QQ’s graph makes these relations explicit and accessible.

\paragraph{Interfaces.}
The interfaces cover different points in a typical workflow.
The Python API is meant for analysis scripts, where identifiers need to be normalized before counting, filtering, or joining datasets.
The CLI exposes the same database for shell workflows (such as \texttt{qq~get~<id>}).
Sources can also be updated through the CLI with a single command.
The browser explorer is meant for discoverability: users can search a name or code, check the identifiers and metadata attached to the resulting languoid, explore neighboring nodes in the graph, and find external resources.

\paragraph{Resource discovery and reproducibility.}
Languoids, scripts, and regions can be searched through identifiers and names (both \emph{exonyms} and \emph{endonyms}).
As mentioned, external resources can be attached to a languoid, which allows for easy discovery in external resources. 
In the current database version, languoids are linked to 7,900+ HuggingFace pages, 6,300+ English Wikipedia articles about a particular languoid, 2,500+ WALS, 2,400+ Grambank, and 2,100+ PHOIBLE links for typological information.
New resources can easily be added through providing where a particular identifier is located and what the URL format is.

With QQ, we adhere to and facilitate the FAIR principles\footnote{\href{https://www.go-fair.org/fair-principles/}{go-fair.org/fair-principles/}} of data management \cite{wilkinson2016fair}. The language metadata in QQ is:
\begin{itemize}[itemsep=\enumsepsize,topsep=2pt]
    \item \emph{Findable.} Each release of QQ is deposited on Zenodo under a persistent DOI,\footnote{\href{https://doi.org/10.5281/zenodo.21219618}{doi.org/10.5281/zenodo.21219618}} and its entries are indexed and queryable both through the browser explorer and programmatically via the API through IDs and metadata.

    \item \emph{Accessible.} QQ is retrievable through standard channels, such as by installation via \texttt{pip} from PyPI, cloning via GitHub, or downloading from Zenodo. 
    Because the database ships as a self-contained snapshot, it remains fully accessible offline and is unaffected by outages of upstream sources.
    
    \item \emph{Interoperable.} QQ serializes its database as compressed JSON with a typed schema. Each entry cross-references identifier schemes, e.g., ISO 639-3, Glottocodes, BCP-47 tags, and Wikidata QIDs, so QQ functions as an interoperability layer across otherwise disconnected identifier systems.

    \item \emph{Reusable.} 
    QQ is designed for reuse across research workflows (discovery, normalization, and management) through a single unified interface, replacing the ad hoc pattern of consulting multiple sources and writing one-off scripts at each step.
\end{itemize}

\paragraph{Licensing.}
The QQ software is released under Apache~2.0, and the packaged database under CC BY-SA~4.0.
This separation is important since Creative Commons actively discourages code to be licensed under any CC license.\footnote{See the \href{https://creativecommons.org/faq/\#can-i-apply-a-creative-commons-license-to-software}{CC FAQ} and \citet{lambert2024dual}.}
Our compiled database consists of the \emph{primary} sources, as listed in \autoref{app:sources}.
\emph{External} resources remain under their own licenses; QQ only links to external resources, it does not incorporate their contents.

\section{Workflow Evaluation}
Existing tools address parts of working with language metadata, such as tag parsing, looking up standards or resource curation (\autoref{sec:related}).
To our knowledge, QQ is the first toolkit that specifically targets common metadata issues encountered in multilingual NLP.
Therefore, we evaluate QQ through representative workflows, since no benchmark exists for such tasks.
Each workflow states the problem, shows QQ's use, and the result.\footnote{Scripts: \href{https://github.com/WPoelman/qwanqwa/tree/v1.2.0/case-studies}{github.com/WPoelman/qwanqwa/case-studies}}

\subsection{Auditing HuggingFace Language Tags}
\textbf{Problem.}
Suppose we want to count or inspect language coverage in datasets on the HuggingFace Hub \citep{lhoest2021datasets}.
Dataset cards can include \texttt{language:} tags, but these tags mix identifier conventions.
This means we will encounter the same languoid under different tags, including deprecated or non-standard ones.

\begin{figure}[t]
    \centering
    \includegraphics[width=0.78\linewidth]{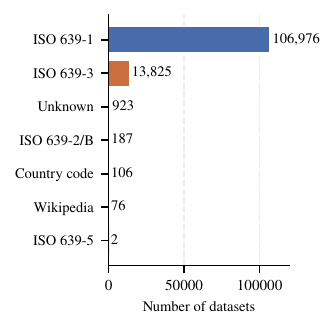}
    \caption{Identifier types observed in HuggingFace datasets after QQ classification.}
    \label{fig:hf-identifier-types}
\end{figure}

\noindent\textbf{Using QQ.}
We collected dataset metadata from the Hub API and QQ classified each unique tag as valid, deprecated,\footnote{Deprecated codes are resolved if a replacement is known.} a country code used as a language code, or unknown.
In June 2026, the Hub API returned 1,051,735 datasets in total, of which 119,583 had at least one \texttt{language:} tag.
Of the 8,237 unique tags, QQ resolves 8,159 tags (99\%) to 7,943 unique languoids.
\autoref{tab:hf-issues} lists identifier issues; \autoref{fig:hf-identifier-types} shows which identifiers are in use.

\noindent\textbf{Result.}
In the absence of QQ, it seems there are 8,159 unique languages on the Hub.
With proper resolving, we see this includes more than 200 duplicates.
Similarly, without QQ, normalization is often handled through local mapping files and manual verification.
QQ provides a systematic alternative by resolving equivalent codes to a single languoid, which helps when counting, filtering, or comparing datasets. 
Previous analyses about language resources, like in \citet{joshi2020state} or \citet{vanesch2024connecting}, benefit from proper normalization to accurately assess language-coverage.
We provide a small, additional analysis about the HF Hub in \autoref{app:sources}, \autoref{fig:hf-coverage}.

\begin{table}[tb]
    \centering
    \footnotesize
    \setlength{\tabcolsep}{3pt}
    \begin{tabularx}{\columnwidth}{@{}l c >{\raggedright\arraybackslash}X@{}}
    \toprule
    Issue & Tags & Examples \\
    \midrule
    Deprecated & 34 & \texttt{eml}, \texttt{ajp}, \texttt{iw} \\
    Country code & 26 & \texttt{gr}, \texttt{cz}, \texttt{jp} \\
    Unknown & 50 & \texttt{multilingual}, \texttt{code}, \texttt{qaa} \\
    \bottomrule
    \end{tabularx}
    \caption{Issues found in the HF Hub audit.}
    \label{tab:hf-issues}
\end{table}

\subsection{Linking Resources}
\begin{figure}
    \centering
    \includegraphics[width=\linewidth]{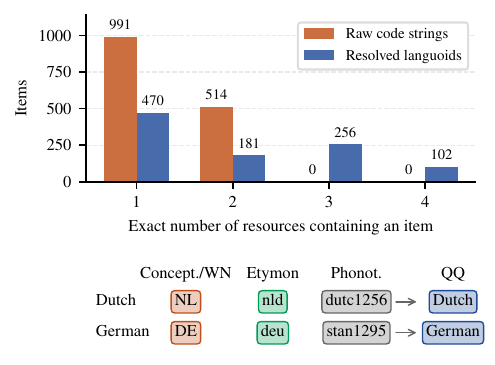}
    \caption{Linking datasets through resolved languoids.}
    \label{fig:linking}
\end{figure}

\textbf{Problem.}
A second recurring task is linking resources.
Suppose we are evaluating on many different datasets that each use a different identifier convention.
Combining these resources is required in evaluation, reporting, exploration, etc.
We use lexical resources as an example because they are a good stress test: they describe overlapping sets of languoids using different identifier conventions.
We link four resources used in lexical and concept-based analysis: Concepticon BabelNet synsets \citep{navigli2010babelnet,list2016concepticon,concepticon}, WordNet synsets~\citep{miller-1994-wordnet}, Etymon~\citep{de-melo-2014-etymological}, and Phonotacticon~\citep{JooHsu+2025+405+431}.
These resources use different identifiers, including ISO 639-2 codes, ISO 639-3 codes, ISO 639-5 family codes, Glottocodes, and a mixture of invalid codes.
Just using exact identifiers hides the overlap between the datasets.
\autoref{fig:linking} shows the overlapping sets between the datasets.

\noindent\textbf{Using QQ.}
The normalization step of QQ resolves different identifiers to a single languoid.
In the current snapshot, QQ resolves 592/600 Concepticon codes, 514/520 WordNet codes, 396/397 Etymon codes, and 517/519 Phonotacticon codes.
Unresolved codes do not belong to any standard.
After resolving, 102 languoids are covered by all four resources.

\noindent\textbf{Result.}
After resolving, datasets can easily be linked through their shared QQ languoids.
This takes away a common hurdle of different identifiers and allows for better interoperability.
Reporting which languages were actually used is also straightforward.
QQ reduces dataset-specific cleanup and makes the process less error-prone than manual cross-referencing.

\subsection{Reporting and Resource Discovery}
\textbf{Problem.}
A third recurring task is reporting which languages were used in a study.
This happens when writing a paper about multilingual NLP, preparing a benchmark, or sanity-checking a language sample before analysis.
The work usually combines normalization, metadata lookup, and manual browsing, often with a few intermediate \textit{ad hoc} corrections.

\begin{table*}[t]
    \centering
    \small
    \setlength{\tabcolsep}{3.5pt}
    \begin{tabularx}{\textwidth}{lcccccccccX}
\toprule
Tool / resource
& \makecell{ID\\norm.}
& \makecell{Retired\\IDs}
& Metadata
& Relations
& \makecell{Ext.\\Links}
& Python
& CLI
& Web
& Snapshot
& Role \\
\midrule
\href{https://www.ethnologue.com/}{Ethnologue}
& \notsure & \notsure & \yes & \yes & \no & \no & \no & \yes & \notsure
& Closed catalogue. \\

\href{https://glottolog.org/}{Glottolog}
& \partially & \no & \yes & \yes & \no & \partially & \no & \yes & \yes
& Classification. \\

\href{https://github.com/google-research/url-nlp/tree/main/linguameta}{LinguaMeta}
& \partially & \partially & \yes & \partially & \no & \no & \no & \no & \yes
& Curated metadata. \\

\href{https://github.com/cisnlp/GlotScript}{Glotscript}
& \no & \no & \partially & \no & \no & \yes & \no & \no & \yes
& Scripts. \\

\href{https://github.com/rspeer/langcodes}{\texttt{langcodes}}
& \yes & \yes & \partially & \no & \no & \yes & \no & \no & \yes
& BCP-47 tags. \\

\href{https://github.com/pycountry/pycountry}{\texttt{pycountry}}
& \partially & \partially & \partially & \no & \no & \yes & \no & \no & \yes
& ISO lists. \\

\href{https://bitbucket.org/robvanderg/distals/}{DistaLs}
& \no & \no & \partially & \no & \no & \yes & \yes & \yes & \yes
& Distances. \\

\href{https://cldf.clld.org/}{CLDF}
& \no & \no & \partially & \partially & \no & \partially & \no & \partially & \partially
& Interchange format. \\

\midrule
QQ
& \yes & \yes & \yes & \yes & \yes & \yes & \yes & \yes & \yes
& Metadata management. \\
\bottomrule
\end{tabularx}

    \caption{Related tools cover adjacent pieces of the problem; QQ provides easy metadata management and discovery for multilingual NLP. \yes~= yes; \no~= no; \partially~= partially supported; \notsure~= unsure.}
    \label{tab:comparison}
\end{table*}

\noindent\textbf{Using QQ.}
We use QQ to generate a typical LaTeX table containing benchmark results; see \autoref{tab:lang-table}.
The output includes relevant metadata such as names, BCP-47 and ISO 639-3 codes, canonical scripts, speaker counts, and language families.
We have used this workflow in multilingual NLP studies: \citet{poelman2025confounding} analyze a large number of experimental settings and conditions, each with different, but overlapping sets of languages and datasets; QQ made this process much easier. We also used QQ to generate tables with consistent English names for the languages used. In \citet{lent2025weaponization}, QQ is used to find language coverage over datasets using different identifiers. And in \citet{accou2026tipa}, QQ is used to generate results and language coverage tables for hundreds of languages across tasks.

\noindent\textbf{Result.}
Reporting which languages are used becomes much easier with QQ.
Instead of manually consulting Wikipedia, Glottolog, and dataset-specific documentation for metadata, researchers can use QQ to inspect, verify, and report languages.
With QQ, we have a single, accessible tool that can be used throughout the entire research workflow.

\section{Related Tools}\label{sec:related}
\autoref{tab:comparison} summarizes how QQ relates to existing tools and resources.
QQ is not meant as a replacement for them; in fact, QQ is interoperable with them.
QQ facilitates more direct interaction with these resources through easier access, discoverability, and tooling.

QQ addresses a different problem from the sources it builds on.
Glottolog \citep{hammarstrom2024glottolog} provides language classifications, phylogenetic data, and Glottocodes \cite{forkel2022glottocodes}.
LinguaMeta \citep{vanesch2022writing,ritchie2024linguameta} provides curated language metadata, such as endangerment, locales, identifiers, and more for living languages.
It is primarily a reference dataset, consisting of 7,000+ JSON files, not a toolkit.
Glotscript \citep{kargaran2024glotscript} is a system for writing system identification; we use its detailed script metadata.
QQ uses these resources as inputs and integrates their fields into shared Languoid, Script, and Region entities.

Ethnologue\footnote{\href{https://www.ethnologue.com/}{ethnologue.com}} is a metadata source we do not include since it has a proprietary license.

Two Python libraries have overlapping features:
\texttt{langcodes}\footnote{\href{https://github.com/rspeer/langcodes}{github.com/rspeer/langcodes}} validates and canonicalizes BCP-47 language tags, and \texttt{pycountry}\footnote{\href{https://github.com/pycountry/pycountry}{github.com/pycountry/pycountry}} exposes ISO 639 standards.
QQ complements these by resolving identifiers across standards to shared Languoids and attaching metadata, relations, and resources.

Some more specialized tools cover related functions: DistaLs \citep{goot2025distals} is a tool for calculating \emph{distances} between languages (defined as ISO 639-3 codes).
QQ is interoperable with DistaLs: identifier normalization can be done through QQ and the resulting ISO 639-3 codes can be used to calculate distances.
Lastly, CLDF resources \citep[\emph{Cross-Linguistic Data Formats;}][]{forkel2018crosslinguistic} provides an interchange format for linguistic datasets.
QQ incorporates datasets that use the CLDF format like WALS, PHOIBLE, and Grambank.
It is therefore again interoperable; CLDF provides a format for data curation, whereas QQ can be used for interacting with this data.

\section{Conclusion}
We present QQ, a language metadata toolkit for multilingual NLP.
QQ merges open sources into one compiled database and makes the same resolved records available in Python, the CLI, and the browser explorer.
The workflows in this paper show how QQ can be used to audit language identifier usage across datasets, link resources with incompatible identifiers, assist in regular multilingual NLP workflows, and discover related languages by family, script, region, dataset inclusion, and more.
QQ offers practical solutions for common metadata issues in multilingual NLP across identifier resolution, metadata lookup, resource discovery, and reporting.

\section*{Limitations}
QQ depends on the sources it integrates.
If a source is missing information or contains an error, QQ may inherit that issue.
Merging sources can also introduce ambiguity when sources disagree.
In practice, this is rare for essential fields: we encountered two conflicting Glottocodes, which were resolved manually.
QQ provides different ways of dealing with such conflicts, see \autoref{app:merging}.

QQ represents a snapshot of language metadata.
It includes historical entities and deprecated identifiers when they are present in the sources, but it does not model how each metadata field changes over time.
Time-aware modeling of names, scripts, regions, and political boundaries would require a different data model and would go more in the direction of language-change research, rather than metadata management.

The HuggingFace audit uses dataset-card metadata returned by the Hub API.
It does not inspect dataset contents or language columns inside individual datasets.
The results should therefore be interpreted as an audit of Hub metadata, not a complete audit of every dataset.
 
The workflow evaluations are representative rather than exhaustive.
They show how QQ supports multilingual NLP metadata tasks, but they do not cover every interface, resource type, or downstream use case.

\section*{Acknowledgments}
In alphabetical order, we thank Stef Accou, Thomas Bauwens, Artur Kulmizev, Esther Ploeger, and Kushal Tatariya for valuable suggestions and for enduring bugs in the development phase.
We also thank Kaja Dobrovoljc, Daan van Esch, Amanda Kann, Marie-Catherine de Marneffe, Tanja Samardžić, and Agata Savary for their insightful feedback.

This work received support from the CA21167 COST action UniDive, funded by COST (European Cooperation in Science and Technology). 
W.P. is funded by a KU Leuven Bijzonder Onderzoeksfonds C1 project with reference C14/23/096.
Y.C. is funded by the Carlsberg Foundation (project no.~CF21-0454) and the Novo Nordisk Foundation (grant no.~NNF24OC0092972).

\section*{Use of AI}
We used AI assistance for creating the web interface, for reviewing ``human-generated'' code for the main library, and for checking the grammar and readability of the (human-generated) paper. 
AI was not used to generate ideas or create other substantial content in this paper.
All the content has been verified and reviewed by the authors, who also take full responsibility. 

\bibliography{clean-bib}
\clearpage

\appendix

\section{Additional Figures}\label{app:sources}
\begin{table*}[ht]
    \onecolumn
    \centering
    \footnotesize
\setlength{\tabcolsep}{4pt}
\renewcommand{\arraystretch}{1.05}

\begin{tabularx}{\textwidth}{>{\raggedright\arraybackslash}p{0.34\textwidth}>{\raggedright\arraybackslash}X>{\raggedright\arraybackslash}p{0.25\textwidth}}
\toprule
Resource & QQ use & License \\
\midrule
\multicolumn{3}{l}{\textbf{Primary Resources}} \\
\midrule
\href{https://glottolog.org/}{Glottolog} \cite{hammarstrom2024glottolog} & Classification, genealogy, Glottocodes. & \mbox{CC BY 4.0} \\
\href{https://github.com/google-research/url-nlp/tree/main/linguameta}{LinguaMeta} \cite{ritchie2024linguameta} & Names, speakers, endangerment, scripts. & \mbox{CC BY-SA 4.0} \\
\href{https://github.com/cisnlp/GlotScript}{Glotscript} \cite{kargaran2024glotscript} & Script metadata. & \mbox{CC BY-SA 4.0} \\
\href{https://iso639-3.sil.org/code_tables/download_tables}{SIL ISO 639-3} & ISO 639-3 identifiers and retired codes. & \mbox{Custom free use with attribution} \\
\href{https://www.loc.gov/standards/iso639-2/}{Library of Congress} & ISO 639-1 and ISO 639-2 mappings. & \mbox{LOC public data / terms} \\
\href{https://www.iana.org/assignments/language-subtag-registry/language-subtag-registry}{IANA} & BCP-47, script, region, and deprecations. & \mbox{Public / Internet Standard} \\
\href{https://www.wikidata.org/}{Wikidata ISO 639-5} & ISO 639-5 family links. & \mbox{CC0} \\
\href{https://www.wikidata.org/}{Wikidata EN sitelinks} & English Wikipedia sitelinks. & \mbox{CC0} \\
\href{https://wikistats.wmcloud.org/}{Wikistats} & Wikipedia edition statistics. & \mbox{CC BY-SA 4.0} \\
\href{https://www.unicode.org/ucd/}{Unicode Character DB} & Script aliases, code-point ranges. & \mbox{Unicode License v3} \\
\midrule
\multicolumn{3}{l}{\textbf{External Resources}} \\
\midrule
\href{https://huggingface.co/datasets}{HF Datasets} \cite{lhoest2021datasets} & Search links; audit. & \mbox{Mixed; Hub terms} \\
\href{https://glottolog.org/}{Glottolog} \cite{hammarstrom2024glottolog} & Reference links. & \mbox{CC BY 4.0} \\
\href{https://www.wikidata.org/}{Wikidata} & Reference links. & \mbox{CC0} \\
\href{https://en.wikipedia.org}{English Wikipedia} & Article links. & \mbox{CC BY-SA 4.0} \\
\href{https://wals.info/}{WALS} \cite{wals} & Typology links. & \mbox{CC BY 4.0} \\
\href{https://grambank.clld.org/}{Grambank} \cite{skirgard2023grambank}  & Typology links. & \mbox{CC BY 4.0} \\
\href{https://phoible.org/}{PHOIBLE} \cite{phoible} & Phonology links. & \mbox{CC BY-SA 3.0} \\
\href{https://universaldependencies.org/}{Universal Dependencies} \cite{nivre2020universal} & Treebank links. & \mbox{UD v2.18 License Agreement} \\
\href{https://afbo.info/}{AfBo} \cite{afbo} & Resource links. & \mbox{CC BY 4.0} \\
\href{https://apics-online.info/}{APiCS} \cite{apics} & Resource links. & \mbox{CC BY 3.0} \\
\href{https://ewave-atlas.org/}{eWAVE} \cite{ewave} & Resource links. & \mbox{CC BY 3.0} \\
\href{https://sails.clld.org/}{SAILS} \cite{sails} & Resource links. & \mbox{CC BY-NC-ND 2.0 DE} \\
\href{https://meta.clld.org/}{CLLD Meta} & Resource links. & \mbox{CC BY 4.0} \\
\bottomrule
\end{tabularx}

    \caption{Primary and external sources used in the construction of QQ. Names are clickable links.}
    \label{tab:sources}
\end{table*}

\begin{figure*}[h]
    \centering
     \begin{subfigure}[b]{0.35\textwidth}
         \centering
         \includegraphics[width=\textwidth]{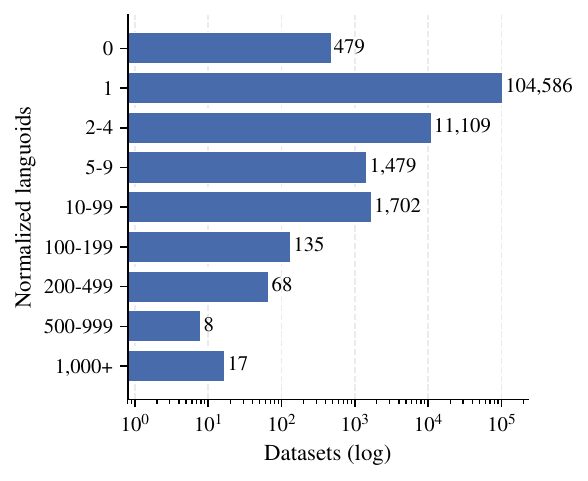}
     \end{subfigure}
     \hspace{2em}
     \begin{subfigure}[b]{0.35\textwidth}
         \centering
         \includegraphics[width=\textwidth]{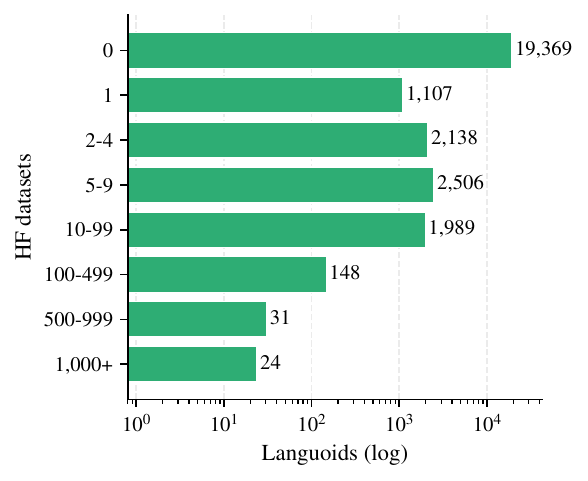}
     \end{subfigure}
     \hfill
    \caption{HuggingFace languoid and dataset coverage. The left plot shows, for each HuggingFace dataset with language metadata, how many unique QQ languoids it contains. Most datasets are monolingual; only a couple are ``massively multilingual.'' The right plot shows, for each QQ languoid, how many HuggingFace datasets contain it. Many languoids appear in only a handful of datasets. Put differently: most datasets cover very few languages, most languages appear in very few datasets or none, and a small number of languages and datasets dominate the multilingual landscape.}
    \label{fig:hf-coverage}
\end{figure*}

\twocolumn

\section{Merging}\label{app:merging}
Since QQ combines many different sources into a single format, we have to choose what to do when sources provide conflicting information.
The most important fields to resolve correctly are the different identifiers.
Luckily, in all the primary sources we merge, we only encountered two conflicting identifiers (a duplicate Glottocode in one of the sources), which was easily resolved.

For other fields, QQ provides two options: source priority or manual resolving.
With priority, each field can specify its preferred source in case of a conflict.
If source $A$ with priority 1 lists the English name for the \texttt{Hant} script as \emph{Han} and source $B$ with priority 2 as \emph{Han (Traditional variant)}, we select source $A$.
With manual selection, the user gets prompted in the command line with all conflicting options for a single field, and can pick which one should be kept.
The results of this operation are stored, so they can be re-used for the next database rebuild.

\begin{table*}[h]
\onecolumn
\footnotesize
\centering
\begin{tabular}{lcccrl}
\toprule
Language & BCP-47 & ISO 639-3 & Script & Speakers & Family \\
\midrule
English & \texttt{en} & \texttt{eng} & Latn & 1.3B & Indo-European \\
French & \texttt{fr} & \texttt{fra} & Latn & 203.2M & Indo-European \\
German & \texttt{de} & \texttt{deu} & Latn & 91.7M & Indo-European \\
Spanish & \texttt{es} & \texttt{spa} & Latn & 468.4M & Indo-European \\
Portuguese & \texttt{pt} & \texttt{por} & Latn & 225.8M & Indo-European \\
Italian & \texttt{it} & \texttt{ita} & Latn & 65.3M & Indo-European \\
Dutch & \texttt{nl} & \texttt{nld} & Latn & 24.1M & Indo-European \\
Russian & \texttt{ru} & \texttt{rus} & Cyrl & 170.2M & Indo-European \\
Mandarin Chinese & \texttt{cmn} & \texttt{cmn} & Hans & 1.3B & Sino-Tibetan \\
Japanese & \texttt{ja} & \texttt{jpn} & Jpan & 119.0M & Japonic \\
Korean & \texttt{ko} & \texttt{kor} & Hang & 75.0M & Koreanic \\
Standard Arabic & \texttt{arb} & \texttt{arb} & Arab & --- & Afro-Asiatic \\
Hindi & \texttt{hi} & \texttt{hin} & Deva & 544.0M & Indo-European \\
Bengali & \texttt{bn} & \texttt{ben} & Beng & 266.0M & Indo-European \\
Tamil & \texttt{ta} & \texttt{tam} & Taml & 81.5M & Dravidian \\
Telugu & \texttt{te} & \texttt{tel} & Telu & 95.0M & Dravidian \\
Thai & \texttt{th} & \texttt{tha} & Thai & 55.0M & Tai-Kadai \\
Vietnamese & \texttt{vi} & \texttt{vie} & Latn & 85.0M & Austroasiatic \\
Standard Indonesian & \texttt{id} & \texttt{ind} & Latn & 171.0M & Austronesian \\
Standard Malay & \texttt{zsm} & \texttt{zsm} & Latn & --- & Austronesian \\
Swahili & \texttt{swh} & \texttt{swh} & Latn & --- & Atlantic-Congo \\
Yoruba & \texttt{yo} & \texttt{yor} & Latn & 28.0M & Atlantic-Congo \\
Igbo & \texttt{ig} & \texttt{ibo} & Latn & 28.0M & Atlantic-Congo \\
Hausa & \texttt{ha} & \texttt{hau} & Latn & 28.2M & Afro-Asiatic \\
Amharic & \texttt{am} & \texttt{amh} & Ethi & 36.0M & Afro-Asiatic \\
Burmese & \texttt{my} & \texttt{mya} & Mymr & 36.0M & Sino-Tibetan \\
Central Khmer & \texttt{km} & \texttt{khm} & Khmr & 15.0M & Austroasiatic \\
Lao & \texttt{lo} & \texttt{lao} & Laoo & 5.1M & Tai-Kadai \\
Georgian & \texttt{ka} & \texttt{kat} & Geor & 3.4M & Kartvelian \\
Eastern Armenian & \texttt{hy} & \texttt{hye} & Armn & 3.0M & Indo-European \\
\bottomrule
\end{tabular}

\caption{Language metadata table generated with the help of QQ (\href{https://github.com/WPoelman/qwanqwa/tree/main/case-studies/latex-tables}{\texttt{source-script}}).}
\label{tab:lang-table}
\end{table*}

\end{document}